\documentclass{article}
\usepackage{spconf,amsmath,graphicx}

\usepackage{enumitem}
\usepackage[dvipsnames]{xcolor}
\usepackage{graphicx}
\setlist{nosep, leftmargin=14pt}
\usepackage{float}
\usepackage{siunitx}

\def\L{{\cal L}}

\title{Leveraging whole slide difficulty in Multiple Instance Learning to improve prostate cancer grading}

\name{Marie Arrivat$^{\star \dagger}$ \qquad Rémy Peyret$^{\star}$ \qquad Elsa Angelini$^{\dagger}$ \qquad Pietro Gori$^{\dagger}$}

\address{$^{\star}$ Primaa, Paris, France \\
    $^{\dagger}$ LTCI, Telecom Paris, Institut Polytechnique de Paris, France}

\begin{document}
%\ninept
%
\maketitle
\begin{abstract}
Multiple Instance Learning (MIL) has been widely applied in histopathology to classify Whole Slide Images (WSIs) with slide-level diagnoses. While the ground truth is established by expert pathologists, the slides can be difficult to diagnose for non-experts and lead to disagreements between the annotators. In this paper, we introduce the notion of Whole Slide Difficulty (WSD), based on the disagreement between an expert and a non-expert pathologist. We propose two different methods to leverage WSD, a multi-task approach and a weighted classification loss approach, and we apply them to Gleason grading of prostate cancer slides. Results show that integrating WSD during training consistently improves the classification performance across different feature encoders and MIL methods, particularly for higher Gleason grades (\textit{i.e.,} worse diagnosis).

\end{abstract}
\section{Introduction}
\label{sec:intro}
Prostate Cancer is one of the most frequent cancers among men. The examination of Whole Slide Images (WSIs) of prostate biopsies by pathologists and their Gleason grading is crucial for orienting the treatment of patients.

Deep Learning applied to histopathology has been very successful in cancer diagnosis and detection, especially for fully supervised tasks, such as patch classification or segmentation. However, gathering both qualitative and quantitative patch-level or pixel-level labels on WSIs is time-consuming and expensive due to the size of the slides (at least 1Go). Hence, the development of weakly supervised methods such as multiple instance learning (MIL) based on slide-level labels has become the new standard. Moreover, the emergence of histopathology-specific foundation models has simplified the extraction of % allows networks to learn directly from 
relevant features. Recent literature \cite{mammadov_self-supervision_2025, shao_multiple_2025} encourages to combine the feature extraction of these foundation models with simple MIL architectures, such as ABMIL \cite{ilse_attention-based_2018}.
When assigning a diagnosis to a slide, some images are more difficult to label than others due to factors such as misleading patterns, small regions of interest or tissue alteration. Most of the time, these elements are detected by expert pathologists but can be missed or mistaken by less qualified pathologists. When comparing diagnoses, easy examples usually lead to consensus among annotators while disagreements can showcase a difficulty inherent to the slide. Thus, gathering a second opinion from a less expert annotator and comparing it to the one of an expert annotator, considered as ground truth, can provide information about the sample difficulty, which can be used %as a prior 
during training. 

In this study, inspired by Query-by-Committee \cite{query}, %Hence, 
we introduce the notion of Whole Slide Difficulty (WSD) that captures the difficulty of a WSI based on the \textit{disagreement} between an expert pathologist and a non-expert one. 
%In this study, we leverage diagnoses provided by an expert pathologist and by a non-expert senior pathologist and infer the WSD based on their disagreement that we use during training.
We consider the labels provided by the expert as ground truth, thus differing from methods exploiting noisy annotations, such as % from noisy annotation leveraging 
\cite{karimi_deep_2020, zhang_multiple_2025}. Furthermore, we distinguish the notion of difficulty of a slide from a lack of certainty or confidence in the labels, as explored in \cite{northcutt_confident_2021, del_amor_labeling_2023}, since experts can be very confident in their diagnosis yet consider the slide to be complex for a non-expert. Finally, our framework differs from multiple annotators (or crowdsourcing ) \cite{herde_multi-annotator_2023,lopez-perez_crowdgleason_2024,ruppli_decoupled_2023} and multi-teacher learning \cite{you_learning_2017}, which focus on learning from several annotators (or teachers) with low or varying experience, without an actual expert considered as ground truth.  % sources, most often creating new labels from labels by mixing several possible truths \cite{ren_label_2023, xu_leveraging_2024, ruppli_decoupled_2023} 

Our contributions are the following. First, we introduce WSD and propose a way to infer it from the disagreement between an expert and non-expert pathologist. Then, we explore methods to leverage WSD as a prior during MIL training for histopathology.

% % \begin{itemize}
% %     \item Histopathology + prostate cancer intro
% %     \item MIL intro + background
% %     \item Context of slide difficulty \& disagreement
% %     \item Differenciation with the other paradigms
% %     \item Intro of the proposed methods and contributions of the paper
% % \end{itemize}

% \begin{figure}[htb]
% \begin{minipage}[b]{1.0\linewidth}
%   \centering
%   \centerline{\includegraphics[width=8.5cm]{ex_image.png}}
% %  \vspace{2.0cm}
% \end{minipage}
% \caption{Example of difficult pattern.}
% \label{fig:res}
% %
% \end{figure}

% \section{Related Work}
% \label{sec:relatedwork}

% \section{Motivation}
% \label{sec:motivation}

\section{Methods}
\label{sec:methodology}

% \subsection{Data structure and pre-processing}
% \label{ssec:data}

\paragraph*{Gleason Grades}
When inspecting prostate slides, pathologists look for Gleason patterns, also called Gleason grades of the prostate glands. Tumourous glands degenerate from Gleason Grade 3 to Gleason Grade 5, while Gleason Grades 1 and 2 are benign. From these observations, pathologists assign a Gleason Score to the slide, which consists of the pair of the most frequently occuring grade and of the highest secondary grade, e.g., 3+5 \cite{epstein_prostate_2018}. For prostate cancer, most papers focus on Gleason grading, essentially for tumour segmentation \cite{bulten_artificial_2022, bulten_automated_2020}. Some MIL methods explore the Gleason Group, based on the total sum of the Gleason Score (e.g., 3+5=8) and others the Gleason Score \cite{ren_enhanced_2025, silva-rodriguez_proportion_2022}. We decided to use the highest grade present in the slide, corresponding to the maximum of the two terms of the Gleason Score. This fits well with the MIL paradigm in which one gland of the highest grade is sufficient to label the whole slide. We used 4 classes for classification: Benign, Gleason 3, 4 and 5.

\paragraph*{Dataset}
%\label{sssec:dataset}
We use a private dataset composed of 2,914 HE-stained WSIs along with their Gleason Score. These scores were assigned by a an expert uro-pathologist (ground truth) who had access to all the clinical details of the patients as well as immunohistochemical stainings of the slides. The WSIs were then reviewed by a senior pathologist, \textit{not expert} of the prostate, who gave them a Gleason Score based on the slides only. 1,995 slides were used for training, 507 for validation and 412 for testing, with a similar balance of classes among the sets.
%and no common patients.

\paragraph*{Whole Slide Difficulty}
%\label{sssec:difficultydef}
We introduce a Whole Slide Difficulty (WSD) score which corresponds to the difficulty inherent to a slide and that can be captured by the disagreement between the expert and the non-expert pathologist. In order to infer the WSD, we first define three levels of disagreement between the pathologists:
\begin{itemize}
    \item a \textbf{homogeneous consensus}, for which both pathologists agree on the two terms of the Gleason Score, allowing them to disagree on the order (\textit{e.g.}, 3+4 and 4+3).
    \item a \textbf{heterogeneous consensus}, for which both pathologists agree on the worst Gleason grade but disagree on the other grade of the Gleason Score (\textit{e.g.}, 4+4 and 3+4).
    \item \textbf{no consensus}, for which both pathologists disagree on the worst grade present in the slide (\textit{e.g.}, 4+5 and 4+3).
\end{itemize}

\noindent Slides with less consensus are then considered more difficult than others. In our dataset, 67.7\% of the WSIs have a homogeneous consensus, 14\% have a heterogeneous consensus and 18.3\% have no consensus. Classes with the least consensus correspond to Grade 3 and Grade 5, which tend to be the most complicated patterns to evaluate for non-experts.

%\subsection{Feature extractor and Multiple Instance Learning backbone}
\paragraph*{Architecture backbones}
\label{sssec:mil}
The WSIs are split into non-overlapping patches of dimension 224x224 at magnification x5 that are extracted from a tissue mask, obtained with an in-house trained U-Net. Patches are then passed to histopathological foundation models and their embeddings are used to train MIL architectures. At every epoch, slides are shuffled and all their instances are used (due to the variability of tissue size between the slides, the number of patches ranges from 68 to 1,187). Two foundation models are used: CTransPath \cite{wang_transformer-based_2022} and UNI2-h \cite{chen_towards_2024}. These models have been shown to be robust, relatively small (less than 100M parameters) and can thus meet usual computational limitations of pathologist consoles (i.e., no or small GPUs).
%\cite{filiot_distilling_2025} .
We use five highly-used MIL methods: MaxMIL, which is instance-based, and ABMIL \cite{ilse_attention-based_2018}, CLAM \cite{lu_data_2021}, DSMIL \cite{li_dual-stream_2021} and TransMIL \cite{shao_transmil_2021}, which are embedding-based and which all use an attention mechanism that can be exploited for visualisation and interpretation.
%For all methods, the instance-based MaxMIL that applies a max-pooling of the patches classifications and the embedding-based MIL models ABMIL \cite{ilse_attention-based_2018}, CLAM \cite{lu_data_2020}, DSMIL \cite{li_dual-stream_2021} and TransMIL \cite{shao_transmil_2021} that leverage different strategies for aggregating the embeddings of the patches were used as MIL backbones. 
For each foundation model and MIL method, a baseline slide classifier is first trained using only the annotations of the expert pathologist. Then, we also train two new WSD-based methods that we describe in the following.
% \begin{itemize}
%     \item Patches passed to histopathology-specific foundation models for feature extraction
%     \item Tested 2 feature extractors: CTranspath and UNIv2
%     \item Then, passed to a MIL backbone for the different setups
%     \item Tested 5 MIL backbones: ABMIL, MaxMIL, DSMIL, TransMIL and CLAM
%     \item For each setup, comparison with the classification pipeline only as baseline
% \end{itemize}

\paragraph*{Multi-task learning (MT)}
\label{ssec:multitask}
%Difficulty scores can be inferred from the disagreement levels of slides.
% When fed to classifiers or regressor MIL frameworks, models learn from this data.
%Thus, the first proposed method is a multi-task learning framework. 
The first proposed method is to add a regression head to the baseline classification framework, so that the model can predict the WSD score along with the highest Gleason grade of the slides. Both the regression and the classification heads are trained using a common multi-task loss, defined in Eq. \eqref{eq:loss_multitask}.
% The first proposed method is to use a multi-task learning framework. By adding a regression head to the baseline classification framework, the model can predict a difficulty score, assigned to each slide based on its disagreement level, along with the highest Gleason grade. Both the regression and the classification heads are trained using a common multi-task loss, defined in the equation \eqref{eq:loss_multitask}.
\begin{equation}\label{eq:loss_multitask}
    \L_{\text{MT}} = \alpha \L_{\text{class}} + \beta \L_{\text{reg}}
\end{equation}
% \begin{itemize}
%     \item Regression of the difficulty level (tried on its own to validate the method)
%     \item Regression and classification heads with multi-task loss
% \end{itemize}
\paragraph*{Classification loss weight}
\label{ssec:lossweight}
The second proposed method is a weighting of the classification loss by a difficulty weight $w_{\text{WSD}}$ assigned to the slides based on their disagreement level. Empirically, we found that %For instance, 
slides with homogeneous consensus should receive a weight $w_{\text{HoC}}$ of 1.0, slides with heterogeneous consensus a weight $w_{\text{HeC}}$ between 1.3 and 4.0 and slides with no consensus a weight $w_{\text{NC}}$ between 2.0 and 10.0, depending on the used foundation model and MIL architecture, to promote difficult slides. The resulting loss used for training is defined in Eq. \eqref{eq:loss_weighted} and \eqref{eq:weights}.

% \begin{equation}\label{eq:loss_weighted}
% \L_{\text{weighted}} = w_{\text{WSD}}\L_{\text{class}}
% \quad 
% w_{\text{WSD}} =
% \begin{cases}
% w_{\text{NC}} & \text{no consensus,} \\
% w_{\text{HeC}} & \text{heterogeneous consensus,} \\
% w_{\text{HoC}} & \text{homogeneous consensus.}
% \end{cases}
% \end{equation}

\begin{equation}\label{eq:loss_weighted}
\L_{\text{weighted}} = w_{\text{WSD}}\L_{\text{class}}
\end{equation}

\begin{equation}\label{eq:weights}
w_{\text{WSD}} =
\begin{cases}
w_{\text{NC}} & \text{no consensus,} \\
w_{\text{HeC}} & \text{heterogeneous consensus,} \\
w_{\text{HoC}} & \text{homogeneous consensus.}
\end{cases}
\end{equation}

\paragraph*{Hyperparameters and Experimental setup}
In all experiments, Cross-Entropy (CE) is used for $\L_{\text{class}}$ and Mean Squared Error (MSE) for $\L_{\text{reg}}$. Optimal hyper-parameters, that is $(\alpha, \beta)$ for the multi-task approach, classification weights $w_{\text{WSD}}$ for the second approach and learning rates for each architecture are empirically found based on grid searches on the validation set. 
% For the weighted loss approach, several weightings of the classification were tested, giving more weight to difficult slides, more weight to easy slides, giving same or different weights to slides with no or heterogeneous consensus.
% For the teacher-student approach, different temperatures were tested in the Kullback–Leibler divergence.
% Grid searches were performed for all these hyperparameters as well as the learning rate for each architecture. Adam optimiser was used.
%Multi-task learning and classification loss weighting were evaluated on two feature extractors and five MIL backbones and the teacher-student approach on ABMIL and TransMIL using CTransPath, the two best models overall.
All experiments are ran using Adam optimizer and each one is ran twice using two random seeds. The average value on the test set is reported. We assess statistical significance using a paired permutation test and estimate confidence intervals via bootstrap.

\section{Results and discussion}
\label{sec:results}

\begin{table*}[!t]
\caption{Results of different methods using CTransPath as feature extractor and five MIL backbones. Values are mean over two random seeds (95\% CI), where $*$ indicates p-value $<$ 0.05. Improvements compared to the baseline are highlighted in bold.}
\label{tab:results-ctranspath}
\resizebox{\textwidth}{!}{%
\begin{tabular}{l|cc||cc|cc|cc|cc|}
\cline{2-11}
 &
  \multicolumn{2}{c||}{\textbf{MaxMIL}} &
  \multicolumn{2}{c|}{\textbf{ABMIL}} &
  \multicolumn{2}{c|}{\textbf{CLAM}} &
  \multicolumn{2}{c|}{\textbf{DSMIL}} &
  \multicolumn{2}{c|}{\textbf{TransMIL}} \\ \hline
\multicolumn{1}{|l|}{\textbf{Methods}} &
  \multicolumn{1}{c|}{\textbf{Bal. Acc.}} &
  \textbf{W. F1-Score} &
  \multicolumn{1}{c|}{\textbf{Bal. Acc.}} &
  \textbf{W. F1-Score} &
  \multicolumn{1}{c|}{\textbf{Bal. Acc.}} &
  \textbf{W. F1-Score} &
  \multicolumn{1}{c|}{\textbf{Bal. Acc.}} &
  \textbf{W. F1-Score} &
  \multicolumn{1}{c|}{\textbf{Bal. Acc.}} &
  \textbf{W. F1-Score} \\ \hline
\multicolumn{1}{|l|}{Classification baseline} &
  \multicolumn{1}{c|}{68.3} &
  75.4 &
  \multicolumn{1}{c|}{71.5} &
  79.7 &
  \multicolumn{1}{c|}{72.1} &
  79.8 &
  \multicolumn{1}{c|}{70.4} &
  78.4 &
  \multicolumn{1}{c|}{73.0} &
  81.6 \\ \hline
\multicolumn{1}{|l|}{Multi-task learning} &
  \multicolumn{1}{c|}{67.8 (-0.7, +0.7)} &
  75.2 (-0.7, +0.6) &
  \multicolumn{1}{c|}{\textbf{73.2} (-1.5, +1.4)} &
  \textbf{80.1} (-1.7, +1.5) &
  \multicolumn{1}{c|}{70.2 (-2.4, +2.4)} &
  78.7 (-2.3, +2.5) &
  \multicolumn{1}{c|}{\textbf{72.1} (-2.5, +2.3)} &
  \textbf{79.7} (-2.6, +2.3) &
  \multicolumn{1}{c|}{70.5 (-1.0, +0.8)} &
  \textbf{81.7} (-1.1, +0.8) \\ \hline
\multicolumn{1}{|l|}{Weighted classification loss} &
  \multicolumn{1}{c|}{\textbf{68.5} (-1.7, +1.4)} &
  75.0 (-1.8, +1.3) &
  \multicolumn{1}{c|}{\textbf{75.6} (-1.4, +1.9) $*$} &
  \textbf{81.6} (-1.4, +2.1) &
  \multicolumn{1}{c|}{\textbf{74.6} (-1.9, +1.9) $*$} &
  \textbf{80.0} (-1.9, +2.0) $*$ &
  \multicolumn{1}{c|}{69.7 (-2.0, +1.8)} &
  78.1 (-2.0, +1.8) &
  \multicolumn{1}{c|}{\textbf{75.8} (-1.1, +1.4) $*$} &
  \textbf{82.1} (-1.0, +1.6) $*$ \\ \hline
\end{tabular}%
}
\end{table*}

\begin{table*}[!t]
\caption{Results of different methods using UNI2-h as feature extractor and five MIL backbones. Values are mean over two random seeds (95\% CI), where $*$ indicates p-value $<$ 0.05. Improvements compared to the baseline are highlighted in bold.}
\label{tab:results-uni}
\resizebox{\textwidth}{!}{%
\begin{tabular}{l|cc||cc|cc|cc|cc|}
\cline{2-11}
 &
  \multicolumn{2}{c||}{\textbf{MaxMIL}} &
  \multicolumn{2}{c|}{\textbf{ABMIL}} &
  \multicolumn{2}{c|}{\textbf{CLAM}} &
  \multicolumn{2}{c|}{\textbf{DSMIL}} &
  \multicolumn{2}{c|}{\textbf{TransMIL}} \\ \hline
\multicolumn{1}{|l|}{\textbf{Methods}} &
  \multicolumn{1}{c|}{\textbf{Bal. Acc.}} &
  \textbf{W. F1-Score} &
  \multicolumn{1}{c|}{\textbf{Bal. Acc.}} &
  \textbf{W. F1-Score} &
  \multicolumn{1}{c|}{\textbf{Bal. Acc.}} &
  \textbf{W. F1-Score} &
  \multicolumn{1}{c|}{\textbf{Bal. Acc.}} &
  \textbf{W. F1-Score} &
  \multicolumn{1}{c|}{\textbf{Bal. Acc.}} &
  \textbf{W. F1-Score} \\ \hline
\multicolumn{1}{|l|}{Classification baseline} &
  \multicolumn{1}{c|}{65.7} &
  71.7 &
  \multicolumn{1}{c|}{67.3} &
  77.8 &
  \multicolumn{1}{c|}{70.0} &
  80.3 &
  \multicolumn{1}{c|}{67.2} &
  78.9 &
  \multicolumn{1}{c|}{67.4} &
  78.5 \\ \hline
\multicolumn{1}{|l|}{Multi-task learning} &
  \multicolumn{1}{c|}{\textbf{66.5} (-1.0, +1.1)} &
  \textbf{72.2} (-1.0, +1.0) &
  \multicolumn{1}{c|}{\textbf{69.2} (-1.3, +1.5) $*$} &
  \textbf{78.8} (-1.4, +1.7) &
  \multicolumn{1}{c|}{\textbf{70.6} (-1.4, +1.2)} &
  \textbf{80.5} (-1.8, +1.2) &
  \multicolumn{1}{c|}{\textbf{70.4} (-1.5, +1.6)} &
  \textbf{79.5} (-1.5, +1.8) $*$ &
  \multicolumn{1}{c|}{67.3 (-2.0, +1.9)} &
  \textbf{79.1} (-2.2, +2.0) \\ \hline
\multicolumn{1}{|l|}{Weighted classification loss} &
  \multicolumn{1}{c|}{\textbf{66.7} (-1.8, +1.5)} &
  \textbf{73.9} (-1.8, +1.5) &
  \multicolumn{1}{c|}{\textbf{67.4} (-2.6, +2.9)} &
  75.6 (-2.9, +3.2) &
  \multicolumn{1}{c|}{69.7 (-1.3, +1.5)} &
  79.3 (-1.3, +1.7) &
  \multicolumn{1}{c|}{\textbf{69.7} (-1.8, +2.3)} &
  77.1 (-1.8, +2.6) &
  \multicolumn{1}{c|}{\textbf{69.4} (-1.4, +1.5) $*$} &
  \textbf{79.0} (-1.3, +1.6) \\ \hline
\end{tabular}%
}
\end{table*}

\begin{table*}[!t]
\caption{Accuracy per class for the classification baseline and the best WSD-based (WSD-B) method for each MIL backbone, using CTransPath as feature extractor. The best results are highlighted in bold.}
\label{tab:results-accuracy-class-ctranspath}
\resizebox{\textwidth}{!}{%
\begin{tabular}{l|cc||cc|cc|cc|cc|}
\cline{2-11}
 &
  \multicolumn{2}{c||}{\textbf{MaxMIL}} &
  \multicolumn{2}{c|}{\textbf{ABMIL}} &
  \multicolumn{2}{c|}{\textbf{CLAM}} &
  \multicolumn{2}{c|}{\textbf{DSMIL}} &
  \multicolumn{2}{c|}{\textbf{TransMIL}} \\ \hline
\multicolumn{1}{|l|}{\textbf{Class}} &
  \multicolumn{1}{c|}{\textbf{Baseline}} &
  \textbf{WSD-B method} &
  \multicolumn{1}{c|}{\textbf{Baseline}} &
  \textbf{WSD-B method} &
  \multicolumn{1}{c|}{\textbf{Baseline}} &
  \textbf{WSD-B method} &
  \multicolumn{1}{l|}{\textbf{Baseline}} &
  \multicolumn{1}{l|}{\textbf{WSD-B method}} &
  \multicolumn{1}{l|}{\textbf{Baseline}} &
  \multicolumn{1}{l|}{\textbf{WSD-B method}} \\ \hline
\multicolumn{1}{|l|}{Benign} &
  \multicolumn{1}{c|}{\textbf{79.3}} &
  78.2 &
  \multicolumn{1}{c|}{\textbf{91.7}} &
  90.2 &
  \multicolumn{1}{c|}{\textbf{90.9}} &
  90.0 &
  \multicolumn{1}{c|}{89.9} &
  \textbf{92.0} &
  \multicolumn{1}{c|}{\textbf{91.5}} &
  90.0 \\ \hline
\multicolumn{1}{|l|}{Gleason 3} &
  \multicolumn{1}{c|}{\textbf{56.3}} &
  53.1 &
  \multicolumn{1}{c|}{58.3} &
  \textbf{65.6} &
  \multicolumn{1}{c|}{\textbf{62.0}} &
  \textbf{62.0} &
  \multicolumn{1}{c|}{\textbf{59.9}} &
  56.3 &
  \multicolumn{1}{c|}{63.0} &
  \textbf{68.8} \\ \hline
\multicolumn{1}{|l|}{Gleason 4} &
  \multicolumn{1}{c|}{\textbf{88.3}} &
  \textbf{88.3} &
  \multicolumn{1}{c|}{\textbf{83.5}} &
  \textbf{83.5} &
  \multicolumn{1}{c|}{80.1} &
  \textbf{80.6} &
  \multicolumn{1}{c|}{79.2} &
  \textbf{85.0} &
  \multicolumn{1}{c|}{\textbf{87.4}} &
  84.0 \\ \hline
\multicolumn{1}{|l|}{Gleason 5} &
  \multicolumn{1}{c|}{47.4} &
  \textbf{52.6} &
  \multicolumn{1}{c|}{52.6} &
  \textbf{63.2} &
  \multicolumn{1}{c|}{55.3} &
  \textbf{65.8} &
  \multicolumn{1}{c|}{52.6} &
  \textbf{55.3} &
  \multicolumn{1}{c|}{50.0} &
  \textbf{60.5} \\ \hline
\end{tabular}
}
\end{table*}

\begin{figure}[ht]
\caption{Attention maps of ABMIL on CTransPath with the baseline method and with the weighted classification loss. This slide was labeled as 3+3 and had no consensus. The baseline model focuses on irrelevant patches and classifies the slide as Benign while the WSD-based model focuses on a patch containing a Gleason 3 gland (circled in black) and correctly classifies the slide as Gleason 3.}
\label{fig:amapg3}
\begin{minipage}[b]{1.0\linewidth}
  \centering
  \centerline{\includegraphics[width=8.5cm]{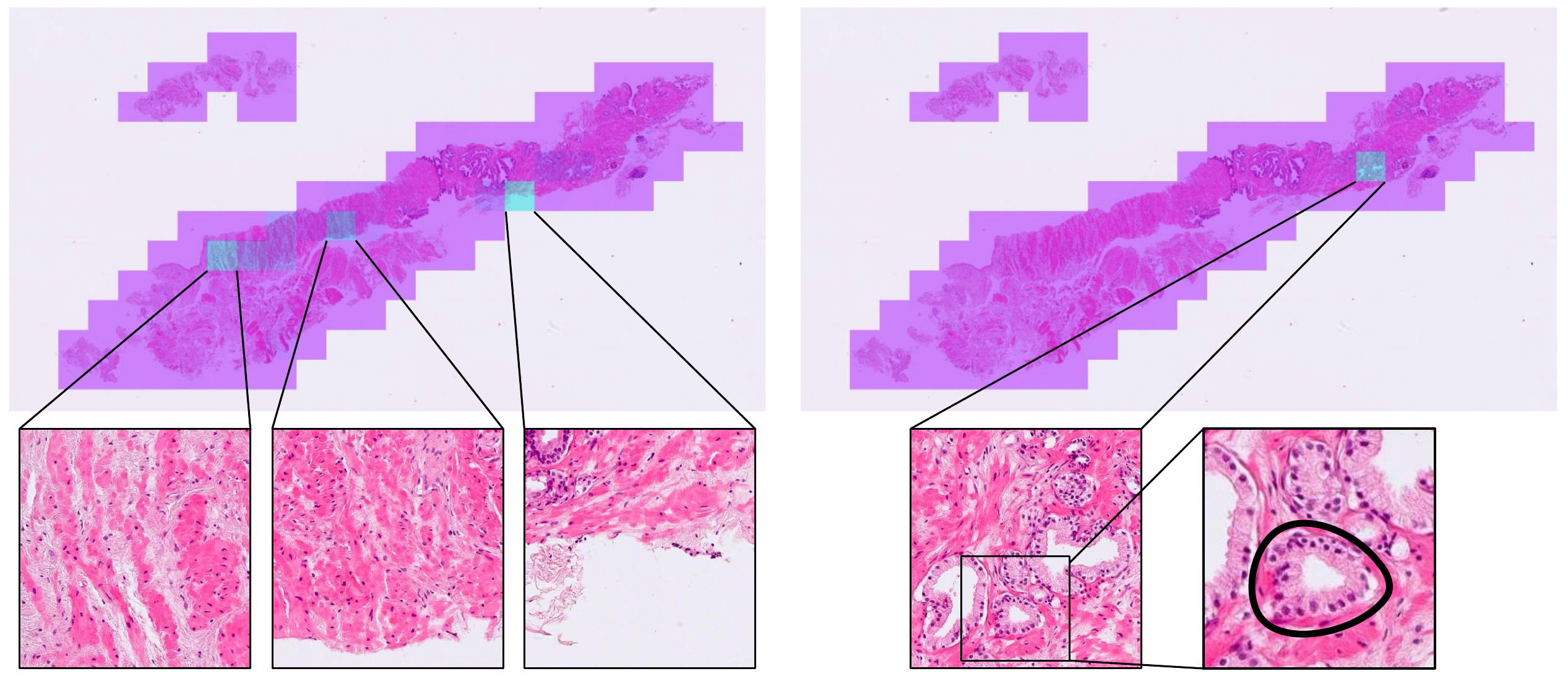}}
%  \vspace{2.0cm}
\end{minipage}
\end{figure}

\begin{table}[ht]
\caption{Impact of ($\alpha$,$\beta$) on multi-task learning using CTransPath and ABMIL.}
\label{tab:ablation-multitask}
\resizebox{\columnwidth}{!}{%
\begin{tabular}{|c|c||c|c|c|c|c|}
\hline
\textbf{($\alpha$,$\beta$)} & (1,0) & (1,1) & (1,2) & (1,3) & \textbf{(1,10)} & (1,50) \\ \hline
\textbf{Bal. Acc.} & 71.5 & 72.9 & 72.6 & 71.7 & \textbf{73.2} & 73.7 \\ \hline
\textbf{W. F1-Score} & 79.7 & 80.0 & 79.7 & 79.9 & \textbf{80.1} & 79.5 \\ \hline
\end{tabular}
}
\end{table}

\begin{table}[ht]
\caption{Impact of WSD on classification loss weighting approach using CTransPath and ABMIL. $w_{\text{NC}}$, $w_{\text{HeC}}$ and $w_{\text{HoC}}$ are the weights given to slides with no consensus, heterogeneous consensus and homogeneous consensus.}
\label{tab:ablation-wloss}
\resizebox{\columnwidth}{!}{%
\begin{tabular}{|c|c||c||c|c|c|c|}
\hline
\textbf{$(w_{\text{NC}}, w_{\text{HeC}}, w_{\text{HoC}})$} & (1,1,1) & (1,1.7,2) & (2,1.3,1) & (4,2,1) & \textbf{(4,3,1)} & (4,4,1) \\ \hline
\textbf{Bal. Acc.} & 71.5 & 71.2 & 75.2 & 75.5 & \textbf{75.6}& 75.6 \\ \hline
\textbf{W. F1-Score} & 79.7 & 78.2 & 81.2 & 81.4 & \textbf{81.6} & 81.5 \\ \hline
\end{tabular}
}
\end{table}

Results of the multi-task learning and the weighted classification loss methods compared with the classification baseline using CTransPath and UNI2-h as feature extractors and five MIL backbones are presented in Table \ref{tab:results-ctranspath} and Table \ref{tab:results-uni}, respectively.
Overall, leveraging difficulty improves performances for Gleason grading. The balanced accuracy is consistently increased by at least one of the WSD-based methods (+2.0 points on average).
%However, the MaxMIL backbone is the least affected by these methods. We explain this by the fact that MaxMIL is an instance-based method and that WSD is a bag-level concept which is more fit for embedding-based architectures.

Table \ref{tab:results-accuracy-class-ctranspath} details the impact of the integration of WSD on each class and shows that the best WSD-based method consistently improves the Gleason 5 accuracy (+7.9 points on average), which is the most critical and difficult pattern to grade.

Figure \ref{fig:amapg3} compares the attention maps of the baseline and of the weighted classification loss method using ABMIL and CTransPath on a difficult 3+3 slide. It shows that integrating WSD helps the model to focus on the relevant patch. 

The impact of the hyperparameters for multi-task learning and classification loss weighting using ABMIL and CTransPath is presented in Tables \ref{tab:ablation-multitask} and \ref{tab:ablation-wloss}. Our analysis suggests that, for multi-task learning,  ($\alpha$, $\beta$) combinations work best when they bring both losses on the same order of magnitude. For classification loss weighting, performance improvements are obtained when promoting difficult slides, while promoting easy slides leads to worse results.
\section{Conclusion}
\label{sec:conclusion}
In this paper, we introduced the notion of Whole Slide Difficulty based on the disagreement between an expert pathologist and a non-expert pathologist. We proposed two different methods to leverage this difficulty as a prior during MIL training for Gleason grading of prostate cancer slides: a multi-task approach and a weighted-loss classification approach. The methods were tested with two foundation models for feature extraction and with five MIL backbones. The results showed that using WSD improves the classification of prostate WSIs, especially for difficult classes. We plan to evaluate more methods to leverage WSD in the future and to apply the developed methods to other organs, such as skin cancer. Finally, it would be interesting to test the robustness of this study by reproducing it with new pairs of diagnoses from different expert and non-expert pathologists.
%\PG{and maybe apply it to other tumors ? clinical cases ? Maybe add limitations ? maybe say that you tried with just one expert and non-epxert and it would be interesting to test the robustness of your method trying with several experts and non-experts to see if it's repredocble.}.

\section{Compliance with ethical standards}
\label{sec:ethics}
% Reporting on compliance with ethical standards is required
% (irrespective of whether ethical approval was needed for the study) in
% the paper. Authors are responsible for correctness of the statements
% provided in the manuscript. Examples of appropriate statements
% include:
% \begin{itemize}
%   \item ``This is a numerical simulation study for which no ethical
%     approval was required.'' 
%   \item ``This research study was conducted retrospectively using
%     human subject data made available in open access by (Source
%     information). Ethical approval was not required as confirmed by
%     the license attached with the open access data.''
%     \item ``This study was performed in line with the principles of
%       the Declaration of Helsinki. Approval was granted by the Ethics
%       Committee of University B (Date.../No. ...).''
% \end{itemize}
The data for this study were derived from \textit{ex vivo} biopsy samples, acquired and anonymised in compliance with the CNIL \textit{Méthodologie de Référence} MR-004 (ref. 2235420). %This framework ensures patient informed consent for the use of their data in research. The computational analysis presented here contributes to the development of a novel \textit{in vitro} diagnostic tool and involved no direct experimentation on human subjects.
% Reporting on real or potential conflicts of interests, or the absence
% thereof, is required in the paper. Authors are responsible for
% correctness of the statements provided in the manuscript. Examples of
% appropriate statements include:
% \begin{itemize}
%   \item ``No funding was received for conducting this study. The
%     authors have no relevant financial or non-financial interests to
%     disclose.'' 
%   \item ``This work was supported by […] (Grant numbers) and
%     […]. Author X has served on advisory boards for Company Y.'' 
%   \item ``Author X is partially funded by Y. Author Z is a Founder and
%     Director for Company C.''
% \end{itemize}
Marie Arrivat and Rémy Peyret are employees at Primaa.
\bibliographystyle{IEEEbib_short}
\bibliography{CSI_1A, ISBI_2026}
%\bibliography{ISBI_2026_all}

%\printbibliography[title={References}]

\end{document}